\documentclass{article}
\usepackage[utf8]{inputenc}
\usepackage{url}
\usepackage{todonotes}
\usepackage{fullpage}
\usepackage[estonian]{babel}
\usepackage{booktabs}
\usepackage{ragged2e}

\title{Närvivõrgu põhise kõnesünteesi arendamine\\{\Large Tehniline raport}}

\author{Liisa Rätsep$^\tau$, Liisi Piits$^\varepsilon$, Hille Pajupuu$^\varepsilon$, Indrek Hein$^\varepsilon$, Mark Fišel$^\tau$\\~\\
$^\tau$ Tartu Ülikool, arvutiteaduse instituut\\
$^\varepsilon$ Eesti Keele Instituut}
\date{6. oktoober 2020}

\begin{document}

\maketitle


\noindent Projekti eesmärgiks oli 
tõsta eestikeelse
kõnesünteesi kvaliteeti ning muuta seda inimkõnele sarnasemaks.
Täpsem tööplaan ning selle tulemused olid järgmised:

\newcommand{\bsct}[1]{\noindent{\textbf{#1}}}

\subsubsection*{Panna kokku andmestik transkribeeritud kõnest mitmelt kõnelejalt}
Täpsem tulemuste kirjeldus on \ref{sctData}. peatükis. Kokkuvõte:

\begin{itemize}
    \item Uudislausete kõnekorpus~\cite{speech_2020}. 65,9 tundi / 36k lauset. 4 kõnelejat (Mari, Albert, Vesta ja Kalev). \\  Allalaadimine (CC-BY-4.0 litsentsiga): 
    \url{https://konekorpus.tartunlp.ai}
    \item Ilukirjanduskorpused \cite{iluk_2020} 26,5 tundi / 16k lauset. 2 kõnelejat (Meelis ja Külli). \\Allalaadimine (CC-BY-4.0 litsentsiga, registreeritud kasutajatele): \url{https://www.eki.ee/litsents/}
\end{itemize}
    
\subsubsection*{Luua tehisnärvivõrgu põhist kõnesünteesi tarkvara}

Täpsem kirjeldus on \ref{sctSoftware}. peatükis. Kokkuvõte:
\begin{itemize}
    \item Kõnesünteesi tarkvara koos kõigi vajalike mudelitega (MIT litsents):\\ \url{https://koodivaramu.eesti.ee/tartunlp/text-to-speech}
    \item Loodud lahenduse analüüs tõi välja parandamist vajavaid aspekte (vt. all).
\end{itemize}

\subsubsection*{Hinnata tulemusi, analüüsida vigu ning võrrelda teiste olemasolevate lahendustega}

Detailsed tulemused on \ref{sctAnalysis}. peatükis. Kokkuvõte:
\begin{itemize}
    \item Kõige kõrgemalt hinnati nii uudiste kui ilukirjanduse lugemise jaoks mõnesid loodud sünteetilisi hääli (eriti "Külli" ja "Mari" hääli), seejärel EKI HTS hääli ning kõige madalamalt -- Google'i kõnesünteesi.
    \item Loodud kõnesünteesi häältel on vaatamata üldisele meeldivusele välja selgitatud rida probleeme, mida esineb teistel häältel vähem: nt. sõnade vahele jätmine, kordamine, liiga järsk heli lõpp jm.
    \item Analüüsi järelduseks on mh see, et suurimatest probleemidest lahti saamiseks peab parendama ja ühtlustama kõneandmete puhastamist ning seejärel treenimist kordama. Samuti kasutati lause tasemel tehtud hindamisel ja analüüsi tegemisel vanemaid HTS mudeleid ning seetõttu said HTS hääled madalama fraseerimise hinnangu.
    \item Google'i kõnesünteesil on peamiseks probleemiks ebaloomulik fraseerimine ja omasõnade vale hääldamine (mida märgati vastavalt 76.5\% ja 60\% analüüsitavatest lausetest).
\end{itemize}

\section{Kõneandmed}
\label{sctData}

Kõneandmed koosnesid eestikeelsetest uudistest (lugenud tudengid, kaks meest ja kaks naist) ja ilukirjandusest (lugenud professionaalne meessoost diktor ja naissoost näitleja).

~ 

\noindent \begin{tabular}{|p{7cm}||p{8.5cm}|}
\hline
\textbf{Uudised} & \textbf{Ilukirjandus} \\
\hline
\hline
\textbf{Mari}: 12 510 lauset / 22,6 tundi & \textbf{Meelis} Kompuse loetud ilukirjandus (Tammsaare\\
\textbf{Albert}: 11 884 lauset / 21 tundi & ``Tõde ja õigus''), 9 959 lauset / 18,5 tundi \\
\textbf{Kalev}: 8 663 lauset / 16,8 tundi & \textbf{Külli} Reinumäe loetud ilukirjandus (pikemad katkendid \\
\textbf{Vesta}: 2 960 lauset / 5,5 tundi & viiest kaasaegsest romaanist), 6 183 lauset / 8 tundi \\
(Nimed muudetud privaatsuse säilitamiseks) & \\
\textbf{Kokku: 36 017 lauset / 65,9 tundi} & \textbf{Kokku: 16 142 lauset / 26,5 tundi} \\
\hline
\end{tabular}

~ 

\noindent Lisaks olid kõnesünteesi mudelite treenimiseks meie kasutuses ka järgmised lisaandmed:
\begin{itemize}
    \item ERRi raadiouudiste korpus (7 kõnelejat, 14,8 tundi, 11 016 lauset),
    \item EKI üksiklausete korpused \cite{yksikl_2020}
    \begin{itemize}
        \item Kersti Kreismann (2 021 lauset, 3,3 tundi),
        \item Liivika Hastin (2 000 lauset, 3,3 tundi),
        \item Külli Reinumägi (1 524 lauset, 2,6 tundi),
        \item Meelis Kompus (1 000 lauset, 2,1 tundi)
    \end{itemize}
\end{itemize}

Seega on terve andmestiku suurus 118.5 tundi kõnet 13 kõnelejaga. Nendest kasutati sünteetiliste häälte loomiseks ainult 6 (uudiste ja ilukirjanduse korpusest), lisaandmeid kasutati mudeli üldiseks tugevdamiseks.

\section{Kõnesünteesi tarkvara}
\label{sctSoftware}

Kõnesünteesi mudeli treenimiseks kasutati eesti keelele kohandatud Deep Voice 3 tarkvara, mis põhineb tehisnärvivõrkudel ja võimaldab mitmehäälse kõnesünteesi mudeli treenimist. Sama lahendust kasutati ka TÜ varasema prototüübi loomisel, kus treeniti mudel uudiste korpuste ja ERRi raadiouudiste korpuse põhjal. Mudelit treeniti TÜ teadusarvutuste keskuse~\cite{https://doi.org/10.23673/ph6n-0144} serverites 45 päeva kasutades Tesla V100 GPU-d ja kõigi kuue kõneleja andmeid.

Et näha, kuidas ilukirjandusliku kõnekorpuse kasutamine samade meetoditega kõnesünteesi kvaliteeti mõjutab, treenisime lisaks erinevate andmehulkadega lisamudeleid kasutades näiteks ainult ilukirjandusel põhinevaid kõnekorpuseid, eemaldades korpustest kõik eeltöötlust vajavad laused, lisades ERRi avaliku raadiouudiste korpuse ja erinevate EKI üksiklausete korpuste andmeid või jagades Meelis Kompuse ilukirjanduskorpuse kolmeks erinevaks hääleks (naistegelase otsekõne, meestegelase otsekõne, jutustaja laused). Nende mudelite kvaliteet oli esialgsel hindamisel märkimisväärselt madalam või lõplikule kuuehäälsele mudelile piisavalt sarnane, et ei õigustanud enda kasutust.

Kood ja treenitud mudel on kättesaadav Deep Voice 3 repositooriumis\footnote{\url{https://github.com/TartuNLP/deepvoice3_pytorch/releases/kratt-v1.0}}. Lisaks on kõnesünteesi API kood koos treenitud mudelitega kättesaadav riiklikus koodivaramus\footnote{\url{https://koodivaramu.eesti.ee/tartunlp/text-to-speech}} ja mudelit saab kasutada TÜ kõnesünteesi veebidemo\footnote{\url{https://www.neurokone.ee}} kaudu, kus on kirjeldatud ka meie avalik API.

Uuendati ka eeltöötlusskripte\footnote{\url{https://github.com/TartuNLP/tts_preprocess_et}}. Eeltöötlusega oli juba tegelenud mitu inimest ja selle üldstruktuur oli varasemast paigas. Algse koodi ülesandepüstitus – katta kõik asendamist vajavad juhtumid u sajas valitud lauses – sai hästi täidetud, seega keskendus koodivärskendus liigse universaalsuse ettevaatlikule tagasitõmbamisele. Suurtähe jadad võivad reaalses tekstis olla akronüümi asemel pealkirjad ja neid ei tule veerida, mitte kõik IVXLCDM jadad pole Rooma nubrid (MM, CV ja DVD), mitte kõik numbrid ei tule kokku võtta (11/12/2020), praktiliselt kõik tüüpilised lühendid on mitmetähenduslikud. Täiendavalt on lisatud tähestikust välja jäävate Unicode’i diakriitiliste märkide taandamine põhitähele.

Kuna uudiste korpus sisaldab lühendite, numbrite ja sümbolitega lauseid, on eeltöötluse kasutamine vajalik ka mudeli treenimisel, seega kasutasime pika treenimisaja tõttu treenimisel projektieelse eeltöötluse versiooni. Hindamisnäidete sünteesimisel oli aga kasutusel eeltöötlusskriptide uuendatud versioonid. 

\section{Hindamine}
\label{sctAnalysis}

Hindamise tulemused koosnevad:
\begin{enumerate}
    \item kõnesünteesi kvaliteedi võrdlusest (\ref{sctComparison}. alampeatükk)
    \item kõnesünteesi detailsemast veaanalüüsist (\ref{sctSubAnalysis}. alampeatükk)
    \item hääle sobivuse hindamisest pikemate sünteesitud kõnelõikude põhjal (\ref{sctSuitable}. alampeatükk)
    \item eeltöötluse hindamisest (\ref{sctPreprocEval}. alampeatükk)
\end{enumerate}

\textbf{Et hinnata treenitud mudeli kvaliteeti} ja võrrelda seda nii teiste eestikeelsete sünteeshäälte kui ka treeningkorpuse häältega, viidi läbi lausete hindamine 20 tudengi seas, kelle seas oli filolooge, kuid kellel puudus varasem kogemus kõnetehnoloogia valdkonnaga. Lausete hindamine põhines teistel tehisnärvivõrkudega kõnesünteesi meetodeid kirjeldavatel teadustöödel \cite{deepvoice3,tacotron2,cotatron}, kus leitakse iga sünteeshääle keskmise arvamuse skoor (MOS). Lisaks Deep Voice 3 mudelitele ja treeningkorpuse näidistele, hinnati ka EKI loodud HTS sünteeshääli\footnote{Üksiklausete hindamisnäidete sünteesimiseks kasutati demolehte \url{https://www.eki.ee/heli/}} ja Google'i tõlkerakenduses\footnote{\url{https://translate.google.com/}} kasutatavat kõnesünteesi.

Hindajad kuulasid 50-lauselisi juhuslikkuse alusel moodustatud ilukirjanduse või uudiste lausete plokke. Kõik hindajad kuulasid plokke erinevas järjekorras ja hindasid lauseid skaalal 1 kuni 5 (0,5 punkti täpsusega), kus 5 on kõige parem hinne. Tudengitel paluti hindamisel lähtuda enda eelistustest olukorras, kus nad kuulavad audioraamatut või uudiseid. Hindamise ajal oli võimalik vaadata ka lausete transkriptsioone. Iga sünteeshääle kohta hinnati vähemalt 25 lauset ilukirjandusest ja 25 lauset uudistest ning iga kõnesünteesi meetodi kohta hinnati vähemalt 100 lauset. Lisaks hinnati iga hääle puhul ka 25 lauset treeningkorpusest, et panna sünteeshäälte meeldivust võrdlusse ka inimhäältega.

Mõlema valdkonna puhul valiti hindamislaused juhuslikult 200-lauselisest valimist ja jälgiti, et sünteeshäälte puhul ei kasutataks lauseid, mis sama mudeli treenimiskorpuses on esinenud. Uudislausete valimist 100 lauset on pärit kõnekorpusest ja 100 lauset WMT News Crawl 2017 uudistekorpusest\footnote{\url{https://www.statmt.org/wmt18/translation-task.html}}. Ilukirjanduslausete valimist 100 lauset pärines samuti kõnekorpustest ja 100 lauset erinevatest eestikeelsetest ilukirjandusteostest.

\textbf{Lisaks lausete hindamisele} paluti tudengitel lausetes esinevaid vigu märgendada. See ülesanne viidi läbi pärast seda, kui kõik laused olid hinnatud, et konkreetset tüüpi vigadele keskendumine ei mõjutaks nende eelistusi ja seeläbi ka hindeid. Märgendamisel toodi välja järgmised vead:
\begin{itemize}
    \item Sõnade vahele jätmine
    \item Sõnade või silpide kordamine või venitamine
    \item Poolik lause
    \item Helitugevuse probleemid
    \item Liiga järsult algav või lõppev lause
    \item Eesti keelele ebaloomulik fraseerimine
    \item Eestikeelsete sõnade vale hääldus (sh probleemid väldete, sõnarõhkude või palatalisatsiooniga)
    \item Võõrsõnade (sh võõrnimede) vale hääldus
    \item Vead sümbolite, numbrite ja lühendite lugemisel
\end{itemize}
Lisaks oli hindajatel võimalik kirjutada kommentaare teiste vigade kohta, mida nad märkasid. Igat lauset märgendas vähemalt kaks inimest.
\textit{}\textit{}

\textbf{Et hinnata sünteeshääle sobivust võimalikesse rakendustesse}  kaasati lisaks projekti käigus loodud Deep Voice 3 sünteeshäältele ka samal meetodil loodud varasemad (kevad 2020) hääleversioonid (edaspidi tähistatud kui \textit{vana}), mille eripäraks oli see, et nad olid treenitud ainult uudistekorpustel ning kasutatud oli vanemat eeltöötlust numbrite ja sümbolite teisendamiseks. Kuigi projekti eesmärk oli hinnata tehisnärvivõrkudega valminud sünteeshääli, et leida sobivaim ilukirjanduse ja uudiste lugemiseks, võeti võrdlusesse ka  peidetud Markovi mudelitel (HMM) HTS-is treenitud hääled kui siiani kõige enam kasutatavad erinevates kõnesünteesirakendustest (eelkõige nende sünteesikiiruse, väikse ressursinõudlikkuse ja\textit{ offline}-võimekuse tõttu). HTS-häälteks valiti   EKI heliraamatute genereerimise rakenduse Vox Populi\footnote{\url{https://heliraamat.eki.ee/voxpopuli}}  Tõnu ja Eva  hääled (treenitud EKI representatiivsel üksiklausete kõnekorpusel,  Tõnul 2874 lauset / 5,5 h  ja  Eval 2020 lauset / 2 h kõnet). Sooviti teada ka seda, kuidas võiks Google’i tõlkerakenduses kasutatav eestikeelne kõnesüntees kuulajate arvates sobida uudiseid ja ilukirjandust lugema.

Sobivuse hindamise testides lähtuti \cite{wagner2017} ja \cite{wagner2019} põhimõtetest. Et kindlaks teha, milliseid sünteeshääli peab potentsiaalne kasutaja sobivaks eri tekstiliike ette lugema, viidi läbi kaks veebipõhist kuulamistesti: Süntesaator uudiste kuulamiseks ja Süntesaator audioraamatu kuulamiseks. Testid ja juhised testijale:
\begin{figure}[h]
    \url{http://peeter.eki.ee:5000/admin/signin}\\
    Nimi: proov2, Salasõna: t3kst1L11k!
\end{figure}

\noindent Testijatel (8 naist  ja 8 meest, vanuses 30-54,  \textit{M}=43,5, \textit{SD} =7,9) tuli kuulata iga sünteeshääle esituses kahte uudist (ilmateade ja majandusuudis) ja kahte katkendit ilukirjandusteosest (ühes dialoogid, teises olukirjeldus), pikkusega umbes 1000 tähemärgi ja hinnata hääle sobivust 7-pallisel Likert-skaalal, kus 1 = ei sobi üldse ... 7 = sobib väga hästi. Testijatel puudus varasem praktiline kogemus eestikeelse kõnesünteesiga, nende hulgas polnud ei filolooge ega kõnetehnolooge.  Arvutati häälele antud hinnangute keskmine uudiste ja ilukirjanduse puhul. 

Et saada ettekujutus hindamise kvaliteedist, arvutati testijate rühmasisene korrelatsioonikoefitsient ICC2k.

\textbf{Eeltöötluse hindamiseks} koostati 177-lauseline kontrollkorpus (varasemale TÜ korpusele lisati veel u 100 lauset), kus numbrite ja lühendite väljakirjutus kontrolliti käsitsi üle.

Kontrollkorpusesse lisati juhuslikke veebilauseid, mis sisaldasid ka keerulisemaid lühendi- ja numbrikombinatsioone: käändelõppudega suurtähelisi lühendeid, mitmetise tähendusega lühendeid, veebiaadresse, mis sisaldasid läbisegi suur- ja väiketähti, telefoninumbreid vms. Enamasti valiti korpusesse ortograafiliselt õigeid lauseid, aga mõnel juhul võeti sisse ka sagedamaid väärkasutusi (nt sidekriips mõttekriipsu asemel vahemike märkimisel, tühikute vale kasutus), kuna ka nende lugemisega võiks sünteeshääl hakkama saada. Korpuses on ka väga keerulisi lauseid, mis vajaks numbrite õigeks käändeks süntaktilist analüüsi.

Hinnang anti lause õigsusele, eraldi anti miinuspunkte juhtude eest, kui lühend on välja loetud millegi muuna (nt km peaks olema käibemaks, aga loetakse kilomeeter) või plusspunkte juhtude eest, kus kontrollkorpus näeb ette lühendi tähthaaval lugemist, aga skript on suutnud õigesti lühendi välja öelda.

\subsection{Kvaliteedi võrdlemise tulemused}
\label{sctComparison}

Üksiklausete hindamise tulemused on esitatud tabelis~\ref{table:mos}. Ootuspäraselt hindasid tudengid kõige kõrgemalt kõnekorpustest pärit lauseid nii uudiste kui ka ilukirjanduse puhul. Inimhäältele järgnesid Deep Voice 3 hääled ning kõige madalamalt hinnati Google'i tõlkemootoris kasutatavat kõnesünteesi. Märkimisväärseid erinevusi üksiklausete hindamisel uue ja vana Deep Voice 3 mudeli vahel ei olnud.

Uudiskorpuse hääli hinnati keskmiselt 1,27 punkti võrra kõrgemalt kui nende põhjal treenitud sünteeshääli, mis on sarnane tulemus ingliskeelse mitmehäälse Deep Voice 3 mudelile~\cite{deepvoice3}, kus vahe oli 1,25 punkti (MOS 4,69 kõnekorpusel ja 3,44 sünteeshäältel). Kõnekorpuse põhjal hinnati parimaks hääleks Vestat ning kõige madalamalt Mari, kuigi Mari oli sünteeshäältest nii mudeli uues kui vanas versioonis hindajate lemmikute seas. Lisaks uudistekorpuse häältele, hinnati uudiste lugejana kõrgelt ka Külli häält.

\begin{table}[ht]
\centering
\caption{Üksiklausete hindamisel põhinevad keskmise arvamuse skoorid uudiste (vasakul) ja ilukirjandustekstide (paremal) lugemisel. Iga skoori kohta on esitatud ka 95\% usaldusvahemik}
\label{table:mos}
\begin{tabular}{lll}
\hline
Hääle tüüp          & Nimi   & MOS  \\ \hline
Kõnekorpus          & Vesta  & $4,65 \pm 0,044$ \\
Kõnekorpus          & Albert & $4,54 \pm 0,050$ \\
Kõnekorpus          & Kalev  & $4,45 \pm 0,052$ \\
Kõnekorpus          & Mari   & $4,37 \pm 0,052$ \\
Deep Voice 3 (vana) & Mari   & $3,62 \pm 0,083$ \\
Deep Voice 3        & Mari   & $3,53 \pm 0,082$ \\
Deep Voice 3        & Külli  & $3,38 \pm 0,080$ \\
Deep Voice 3        & Vesta  & $3,29 \pm 0,074$ \\
Deep Voice 3 (vana) & Albert & $3,26 \pm 0,075$ \\
Deep Voice 3        & Albert & $3,23 \pm 0,074$ \\
Deep Voice 3 (vana) & Vesta  & $3,21 \pm 0,076$ \\
Deep Voice 3 (vana) & Kalev  & $2,90 \pm 0,068$ \\
Deep Voice 3        & Kalev  & $2,86 \pm 0,067$ \\
Deep Voice 3        & Meelis & $2,35 \pm 0,075$ \\
HTS                 & Tõnu   & $2,33 \pm 0,089$ \\
HTS                 & Eva    & $2,25 \pm 0,078$ \\
Google              & -      & $2,10 \pm 0,046$ \\
\end{tabular}
\begin{tabular}{lll}
\hline
Hääle tüüp          & Nimi   & MOS  \\ \hline
Kõnekorpus          & Meelis & $4,89 \pm 0,022$ \\
Kõnekorpus          & Külli  & $4,87 \pm 0,028$ \\
Deep Voice 3 (vana) & Mari   & $3,88 \pm 0,069$ \\ 
Deep Voice 3        & Külli  & $3,66 \pm 0,076$ \\ 
Deep Voice 3        & Mari   & $3,56 \pm 0,081$ \\ 
Deep Voice 3        & Albert & $3,34 \pm 0,072$ \\ 
Deep Voice 3 (vana) & Vesta  & $3,30 \pm 0,070$ \\ 
Deep Voice 3        & Vesta  & $3,25 \pm 0,078$ \\ 
Deep Voice 3 (vana) & Albert & $3,05 \pm 0,080$ \\ 
Deep Voice 3 (vana) & Kalev  & $2,83 \pm 0,070$ \\  
Deep Voice 3        & Kalev  & $2,58 \pm 0,065$ \\  
Deep Voice 3        & Meelis & $2,40 \pm 0,077$ \\  
HTS                 & Tõnu   & $2,12 \pm 0,082$ \\  
HTS                 & Eva    & $2,05 \pm 0,077$ \\  
Google              & -      & $2,04 \pm 0,046$ \\
\\
\\

\end{tabular}
{\textbf{}}\
\justify
\textbf{\textit{Märkus:}} HTS-häälte hindamistulemustesse tuleb suhtuda reservatsiooniga, kuna testi võeti helinäited, mis on sünteesitud fraseerimisveaga. St sünteesimiseks kasutati rakendustes kasutusel mitteolevat versiooni, kus pausi asukoht oli lükkunud kirjavahemärgi või sidesõna kohalt ühe koha võrra edasi. See on tinginud helinäidetes ebaloomuliku intonatsiooni, mis võis mõjutada ka üldist hinnangut.

\end{table}

Ilukirjanduse valdkonnas olid tulemused Külli ja Meelise puhul väga erinevad. Kui Külli sünteeshäält hinnati keskmiselt korpusega võrreldes 1,21 punkti võrra madalamaks, siis Meelise häält 2,49 punkti võrra. Üldiselt eelistati taaskord Mari ja Külli hääli ning tulemuste põhjal pole võimalik väita, et ilukirjanduskorpusega treenitud mudel oleks uudislausete korpusel põhinevate häälte kvaliteeti ilukirjandustekstide lugemisel märgatavalt tõstnud.

\begin{table}[p]
\centering
\caption{Vigade esinemisprotsendid üksiklausetes}
\label{table:annotation}

\begin{tabular}{@{}llccccc@{}}
\toprule
Hääle tüüp & Nimi & \begin{tabular}[c]{@{}c@{}}Sõnade \\ vahele jätmine\end{tabular} & \begin{tabular}[c]{@{}c@{}}Kordamine, \\ venitamine\end{tabular} & \begin{tabular}[c]{@{}c@{}}Poolik\\ lause\end{tabular} & \begin{tabular}[c]{@{}c@{}}Helitugevuse\\  probleemid\end{tabular} & \begin{tabular}[c]{@{}c@{}}Järsk\\ lõpp/algus\end{tabular} \\ \midrule
Kõnekorpus & Mari & 3,6\% & 1,8\% & 0\%& 1,8\% & 14,3\% \\
Kõnekorpus & Kalev & 3,8\% & 1,9\% & 0\%& 1,9\% & 13,2\% \\
Kõnekorpus & Albert & 0\%& 0\%& 0\%& 0\%& 13,3\% \\
Kõnekorpus & Vesta & 0\%& 0\%& 0\%& 3,6\% & 23,2\% \\
Kõnekorpus & Külli & 0\%& 0\%& 0\%& 1,9\% & 1,9\% \\
Kõnekorpus & Meelis & 0\%& 0\%& 0\%& 1,6\% & 0\%\\
\midrule
Deep Voice 3 & Mari & 0,9\% & 21,6\% & 0\%& 29,3\% & 20,9\% \\
Deep Voice 3 & Kalev & 4,3\% & 9,5\% & 0\%& 10,3\% & 13,8\% \\
Deep Voice 3 & Albert & 0,9\% & 7\%& 0\%& 14,8\% & 20\%\\
Deep Voice 3 & Vesta & 0,9\% & 10,5\% & 1,8\% & 12,3\% & 25,4\% \\
Deep Voice 3 & Külli & 2,7\% & 8\%& 1,8\% & 19,5\% & 6,2\% \\
Deep Voice 3 & Meelis & 18,3\% & 14,2\% & 5\%& 15,8\% & 9,2\% \\
\midrule
Deep Voice 3 (vana) & Mari & 0,9\% & 10,3\% & 0,9\% & 17,1\% & 19,7\% \\
Deep Voice 3 (vana) & Kalev & 0\%& 16,7\% & 0,9\% & 14\%& 7\%\\
Deep Voice 3 (vana) & Albert & 0\%& 7,1\% & 4,4\% & 12,4\% & 18,6\% \\
Deep Voice 3 (vana) & Vesta & 0\%& 12,1\% & 0,9\% & 25,9\% & 29,3\% \\
\midrule
HTS & Tõnu & 0\%& 3,5\% & 0\%& 0,9\% & 3,5\% \\
HTS & Eva & 0\%& 1,8\% & 0\%& 2,7\% & 2,7\% \\
\midrule
Google & - & 0\%& 1,3\% & 0\%& 0,4\% & 1,3\% \\
\bottomrule
\end{tabular}
\centering
\begin{tabular}{@{}llcccc@{}}
\\
\toprule
Hääle tüüp & Nimi & 
\multicolumn{1}{c}{\begin{tabular}[c]{@{}c@{}}Ebaloomulik\\ fraseerimine\end{tabular}} & \multicolumn{1}{c}{\begin{tabular}[c]{@{}c@{}}Omasõnade\\ hääldamine\end{tabular}} & \multicolumn{1}{c}{\begin{tabular}[c]{@{}c@{}}Võõrsõnade\\ hääldamine\end{tabular}} & \multicolumn{1}{c}{\begin{tabular}[c]{@{}c@{}}Numbrite, sümbrolite,\\ lühendite vead\end{tabular}} \\ \midrule
Kõnekorpus & Mari & 35,7\% & 10,7\% & 10,7\% & 5,4\% \\
Kõnekorpus & Kalev & 17\%& 9,4\% & 1,9\% & 1,9\% \\
Kõnekorpus & Albert & 18,3\% & 3,3\% & 3,3\% & 5\%\\
Kõnekorpus & Vesta & 10,7\% & 5,4\% & 3,6\% & 0\%\\
Kõnekorpus & Külli & 13\%& 1,9\% & 0\%& 0\%\\
Kõnekorpus & Meelis & 14,5\% & 1,6\% & 0\%& 0\%\\
\midrule
Deep Voice 3 & Mari & 20,7\% & 12,1\% & 17,2\% & 4,3\% \\
Deep Voice 3 & Kalev & 26,7\% & 7,8\% & 12,9\% & 3,4\% \\
Deep Voice 3 & Albert & 28,7\% & 12,2\% & 7,8\% & 1,7\% \\
Deep Voice 3 & Vesta & 14,9\% & 7\%& 7,9\% & 4,4\% \\
Deep Voice 3 & Külli & 29,2\% & 10,6\% & 14,2\% & 3,5\% \\
Deep Voice 3 & Meelis & 40\%& 17,5\% & 16,7\% & 4,2\% \\
\midrule
Deep Voice 3 (vana) & Mari & 22,2\% & 17,9\% & 13,7\% & 5,1\% \\
Deep Voice 3 (vana) & Kalev & 28,1\% & 10,5\% & 12,3\% & 2,6\% \\
Deep Voice 3 (vana) & Albert & 26,5\% & 8,8\% & 8,8\% & 6,2\% \\
Deep Voice 3 (vana) & Vesta & 13,8\% & 6,9\% & 8,6\% & 1,7\% \\
\midrule
HTS & Tõnu & 82,5\% & 12,3\% & 12,3\% & 3,5\% \\
HTS & Eva & 75,7\% & 21,6\% & 12,6\% & 3,6\% \\
\midrule
Google & - & 76,5\% & 60\%& 13,5\% & 4,3\% \\

\bottomrule
\end{tabular}
\justify

\textbf{\textit{Märkus:}} HTS-häälte hindamistulemustesse tuleb suhtuda reservatsiooniga, kuna testi võeti helinäited, mis on sünteesitud fraseerimisveaga. St sünteesimiseks kasutati rakendustes kasutusel mitteolevat versiooni, kus pausi asukoht oli lükkunud kirjavahemärgi või sidesõna kohalt ühe koha võrra edasi.
\end{table}

Et mõista, miks hindajate eelistused kõnekorpuste näidete ja sünteeshäälte puhul erinevad, tuleks andmete sobivust kõnesünteesiks põhjalikumalt analüüsida. On võimalik, et hääletämber või kõnemaneer muudavad mõned kõnelejad närvivõrgupõhiste mudelite treenimiseks sobivamaks või kuulajale meeldivamaks. Näiteks võivad Meelis Kompuse puhul sünteeshääle kvaliteeti mõjutada tekstis esinevad pausid või audioraamatule kohane ilmekus, mida mudel pole võimeline ära õppima. Vesta puhul on treeningkorpuses tema loetud lauseid ka teistega võrreldes märgatavalt vähem, mis võib olla üheks põhjuseks, miks tema sünteeshääle kvaliteeti paremaks ei hinnata. Lisaks eelistatakse peamiselt naishääli ja pigem kõrgemat meeshäält Albertit kui madalama tämbriga Kalevit, mistõttu tuleks uurida, kas Deep Voice 3 või tehisnärvivõrke kasutavad mudelid üldiselt oskavad paremini jäljendada kõrgemaid hääli.

\subsection{Vigade analüüsi tulemused}
\label{sctSubAnalysis}

Üksiklausete märgendamisel esile toodud vigade esinemisprotsendid on esitatud tabelis~\ref{table:annotation}. Nende tulemuste põhjal näeme, et kuigi üksiklausete hindajad eelistasid Deep Voice 3 mudeleid, tõid nad nende puhul välja ka mitmeid probleeme mida HTS ja Google'i mudelites ei esinenud või esines märgatavalt vähem, näiteks sõnade või silpide vahele jätmine, kordamine või lause osaline lugemine. Eriti tugevalt mõjutab sõnade vahele jätmise probleem Meelise sünteeshäält (18,3\% lausetest), kelle puhul oleme täheldanud kalduvust ütlemata jätta lause esimene sõna. Esialgse analüüsi põhjal on tõenäoline, et Meelise korpuses tekitab seda probleemi varieeruva pikkusega vaikus lausete alguses, mida närvivõrk ei oska ette ennustada.

Samuti esines HTS ja Google'i sünteeshäältel vähem probleeme helitugevusega, eriti selle langemisega lause teises pooles, ja liigselt järsu lause alguse või lõpuga, mida olime Deep Voice 3 sünteeshäältel ka juba varem täheldanud. Üllatulikult selgus aga, et viimast probleemi esineb ka uudislausete korpuses ja on sünteeshäälte puhul omane just uudislausete korpuse häältele. Vastupidiselt Meelise korpusele, kus alguses ja lõpus esineb liiga palju vaikust, on uudistekorpusest vaikust varasemate eksperimentide jaoks ära lõigatud ja seda kohati liiga agressiivselt, eriti klusiilidega lõpevate lausete puhul. Seetõttu oleme projekti tulemusena avaldanud uudislausete korpuse eeltöötlemata versiooni ja plaanime järgnevates eksperimentides keskenduda muuhulgas rohkem helifailide eeltöötlusele.

Kõige paremini langeb üksiklausete hindamistulemustega kokku protsent ebaloomuliku fraseerimisega lausetest, kus tudengid tõid peamiselt välja probleeme lause rütmi ja meloodiaga. Kuigi ebaloomuliku fraseerimise hindamise näol on tegemist pigem subjektiivse ülesandega, iseloomustab see hästi hindajate eelistusi. Lisaks on võimalik, et kuna hindamise ajal nägid hindajad ka lausete transkriptsioone, mis aitasid neil lausetest paremini aru saada, eelistati hinnete andmisel eelkõige kõne loomulikkust ja tavapärasest vähem arusaadavust või seda, kui suurt pingutust kuulamine nõuab.

Fraseerimise kategoorias on teiste Deep Voice 3 häältega võrreldes taaskord Meelise häälel kõige kõrgem vea esinemisprotsent. Kõigist sünteeshäältest kõige vähem probleeme fraseerimisega on Vestal, kelle puhul toodi ka kõnekorpuse näidete põhjal kõige vähem fraseerimisprobleeme välja. Samas on hindajad oma kommentaarides Vesta sünteeshäälele ette heitnud monotoonsust ja kvaliteedi langust lause lõpus.

Kõnekorpuste häältest on ebaloomuliku fraseerimise poolest enim välja toodud Mari häält, kuid tundub, et sünteesmudel pole neid ebaloomulikke kõnemaneere omandanud. See on tõenäoliselt ka põhjuseks, miks Mari häält hinnatakse korpuses kõige vähem meeldivaks, kuid sünteeshäälte seas teda pigem eelistatakse. Samuti on Mari korpuse näidete puhul toodud välja probleeme sõnade õige häälduse valimisega. Võõrsõnade ning numbrite, sümbolite ja lühendite veaprotsentide tõlgendamisel tuleks aga silmas pidada, et neid sisaldavate hindamislausete arv võib erinevatel häältel väga erinev olla ning ilukirjanduskorpuste transkriptsioonides olid need juba häälduslikule kujule viidud.

Eestikeelsete sõnade hääldamisega on kõige rohkem raskusi Google'i sünteeshäälel, kus hääldusvigu esineb lausa 60\% lausetest. Google'i puhul on hindajad eraldi välja toonud, et häälel on probleeme õige välte valimisega ja klusiilide hääldamisega. Lisaks on mainitud, et tegemist on ebaloomulikult aeglase rääkijaga.

\subsection{Hääle sobivuse tulemused}
\label{sctSuitable}

Sünteeshäälte sobivus uudiste ja ilukirjanduse lugemiseks on esitatud tabelis~\ref{table:pikad}.
Hindajate käitumine oli konsistentne, rühmasisene korrelatsioonikoefitsient (ICC2k) oli 0,85 (\textit{p} = 0,0001).
\begin{table}[h]
\centering
\caption{Sünteeshääle sobivus uudiste- ja ilukirjanduse lugejaks. \textit{Keskmised hinded 7-pallisel skaalal standardhälbega} }
\label{table:pikad}
\begin{tabular}{llllll}
\hline
\textbf{Sobivus uudisteks}& \textbf{keskmine}& \textbf{sh}& \textbf{Sobivus ilukirjanduseks} &\textbf{keskmine}& \textbf{sh} \\
\hline
Mari (vana) &  5,07 & 1,46 & Külli & 5,34 & 1,31\\
Mari & 4,93 & 1,46 & Vesta (vana) & 5,00 & 1,41\\
Albert (vana) &  4,31 & 1,66 & Mari (vana) & 4,75& 1,37\\
Albert & 4,28&1,71& Mari&4,66&1,47\\
Külli &  4,22 & 1,52 & Vesta & 4,53 & 1,30\\
Vesta & 4,06 & 1,61 & Albert (vana) & 3,97 & 1,40\\
Tõnu&3,47&1,92& Kalev (vana) & 3,55 & 1,71\\
Vesta (vana) & 3,31 & 1,38 & Meelis & 3,34 & 1,70\\
Kalev (vana) & 3,31 & 1,45 &Albert & 3,28 & 1,82\\
Meelis & 3,31 & 2,01 &Kalev & 2,63 & 1,43 \\
Kalev & 3,16 & 1,51 & Eva&2,50&1,69\\
Eva&3,06&1,81&Tõnu&2,44&1,48\\
Google&2,03&1,15&Google&2,38&1,24\\
\hline
\end{tabular}
\end{table}

Uudiste lugemine eeldab neutraalset stiili, ilukirjanduse lugemine märksa ekspressiivsemat stiili, millele aitavad kaasa tähendusrikkad pausid.  Häältest peeti uudiseid lugema sobivaks eelkõige Mari ja Albertit, ent ka ilukirjanduse peal treenitud Küllit. Kuulata uudiseid ilukirjanduslikus lugemisstiilis Külli esituses ei olnud kuulajatele probleem. Siiski oleks ennatlik järeldada, et ka reaalses olukorras kuulataks uudiseid meelsasti ilukirjanduslikus stiilis:  ilukirjanduse peal treenitud Meelise esituses kuulatud uudised nii vastuvõetavad ei olnud. Meelisele antud hinnanguid võisid mõjutada ka tehnilised probleemid, mida kirjeldas eelmine alampeatükk. Projekti raames tehisnärvivõrkudega treenitud häälte ja varasema versiooni häälte vahel märkimisväärset erinevust polnud, v.a Vestal, mille uus versioon sobis kuulajatele vanast rohkem. HTS-häältest peeti uudiste jaoks sobivaks Tõnu. Google'i hääl jäi sobivuselt viimaseks.

Ilukirjanduse lugemiseks peeti kõige sobivamaks Küllit, mis oli treenitud kaasaegse ilukirjanduse korpuse peal. Ent Meelist, mis oli samuti treenitud ilukirjanduskorpuse peal (Tammsaare “Tõde ja õigus“) edastasid mitmed uudiskorpusel treenitud hääled. Tulemused viitavad sellele, et uudistekorpusel treenitud hääl (n-ö neutraalne stiil) ei muutu ilukirjanduse lugemisel kuulaja jaoks sobimatuks.  

Tehisnärvivõrkudega treenitud häältest eelistasid kuulajad ilukirjanduse puhul pigem vana versiooni hääli. HTS-häälete skoor jäi ilukirjanduse puhul alla keskmise.
Naisehääli peeti ilukirjandust lugema sobivamaks, seda ka HTS-häälte puhul.

Arvestada tuleks sellega, et testijatel puudus varasem kokkupuude eestikeelse kõnesünteesiga, mistõttu võivad nende eelistused tegelikus kasutusolukorras olla teistsugused. 
See, mis testijatele meeldis, ei pruugi olla ka koheselt rakendatav, kuid annab aimu, mis suunas edasi töötada.

Kokkuvõttes näitavad tulemused seda, et hääle sobivus sõltub tema rakenduseesmärgist, mistõttu süntees\-häälte hindamisel on alati vaja hindajale öelda, kus seda häält kasutatakse või on kavas kasutama hakata.

\subsection{Eeltöötluse hindamise tulemused}
\label{sctPreprocEval}

Algne eeltöötlus suutis kontrollkorpuse 177-st lausest \textbf{ 87 lauset ehk 49\% } teisendada sarnaselt käsitsi teisendatud korpusega. Lisaks saadi miinuspunkte valesti teisendatud lühendite eest 11 korral ja plusspunkte 4 korral.

Parandatud eeltöötlus suutis kontrollkorpuse 177-st lausest  \textbf {114 lauset ehk 64\% } teisendada sarnaselt käsitsi teisendatud korpusega. Lisaks saadi miinuspunkte valesti teisendatud lühendite eest 6 korral ja plusspunkte ühel korral.

Uuenenud tekstitöölus on paranenud \textbf {15 protsendipunkti} võrra. Miinuseid valesti lahtikirjutatud lühendite eest ligi poole vähem.

\textbf {Lahendatud suuremad probleemid:}
\begin{enumerate}
\item Pikemaid suurtähelisi sõnu ei peeta enam lühenditeks ega veerita tähthaaval.
\item Traditsiooniliselt sõnadena loetavaid suurtähelisi lühendeid (NATO jne) ei veerita tähthaaval. 
\item Väiketähelisi ainult konsonantidest tähekombinatsioone (spp) hääldatakse just tähthaaval. 
\item Suurtäheühendeid (CI, DVD vms) ei loeta enam rooma numbritena. 
\item Vähenenud on lühendite hulk, mis on välja loetud valesti (nt TK, e, a), vt miinusteskoori vähenemist. 
\end{enumerate}
\textbf {Lahendamist vajavad probleemid:}
\begin{enumerate}
\item Numbrid, mida peaks lugema üheliste kaupa (nt isikukood, telefoni- või kontonumber), on välja loetud kümnelisi, sajalisi, tuhandelisi ja miljonilisi sisaldavate arvudena. 
\item Ei oska suurtähelisi lühendeid käänata või käänab valesti kui käändelõpp on küljes. Nt MTÜle (emm-tee-üü), EAS-i (ee-aa-essu).
\item Probleeme esineb keeruliste suur- ja väiketähtede kombinatsioonides, mis esinevad veebiaadressides (https://goo.gl/forms/MrYnlg8Sa4E1Pg5l2) ja nimedes nagu eCoop (ee-tseeoop) ja DigiDoc4 (digi-deeoc neli).
\item Ortograafiliselt mitte päris õigete kriipsu-kombinatsioonide ja tühikute vale kasutuse puhul ei suudeta sobivat varianti välja pakkuda. Nt kui numbrite vahel on mõttekriipsu asemel sidekriipsu kasutatud, siis ei saa tekstitöötlus aru, et tegu vahemikuga, kui koolonile eelneb ja järgneb tühik, siis loeb „jagatud“ isegi kui tegu on suhte näitamisega.
\item Laialdaseks kasutamiseks tuleks võõrnimed transkribeerida häälduspärasele kujule.
\end{enumerate}



\begin{thebibliography}{1}

\bibitem{speech_2020}
M.~Fišel and L.~Rätsep, ``Eesti uudislausete kõnekorpus,'' 2020. \url{https://doi.org/10.15155/9-00-0000-0000-0000-001ABL.}

\bibitem{iluk_2020}
E.~K. Instituut, ``Ilukirjanduse kõnekorpused {M}eelis ja {K}ülli,'' 2020.

\bibitem{yksikl_2020}
E.~K. Instituut, ``{EKI} representatiivsed üksiklausete kõnekorpused,'' 2019,
  2020.

\bibitem{https://doi.org/10.23673/ph6n-0144}
{University of Tartu, HPC}, ``{UT Rocket},'' 2018.

\bibitem{deepvoice3}
W.~Ping, K.~Peng, A.~Gibiansky, S.~{\"{O}}. Arik, A.~Kannan, S.~Narang,
  J.~Raiman, and J.~Miller, ``Deep voice 3: 2000-speaker neural
  text-to-speech,'' {\em CoRR}, vol.~abs/1710.07654, 2017.

\bibitem{tacotron2}
J.~Shen, R.~Pang, R.~J. Weiss, M.~Schuster, N.~Jaitly, Z.~Yang, Z.~Chen,
  Y.~Zhang, Y.~Wang, R.~J. Skerry{-}Ryan, R.~A. Saurous, Y.~Agiomyrgiannakis,
  and Y.~Wu, ``Natural {TTS} synthesis by conditioning wavenet on mel
  spectrogram predictions,'' {\em CoRR}, vol.~abs/1712.05884, 2017.

\bibitem{cotatron}
S.~won Park, D.~young Kim, and M.~chul Joe, ``Cotatron: Transcription-guided
  speech encoder for any-to-many voice conversion without parallel data,''
  2020.

\bibitem{wagner2017}
P.~Wagner and S.~Betz, ``Speech synthesis evaluation: Realizing a social
  turn,'' {\em Tagungsband Elektronische Sprachsignalverarbeitung (ESSV)},
  2017.

\bibitem{wagner2019}
P.~Wagner, J.~Beskow, S.~Betz, J.~Edlund, J.~Gustafson, G.~E. Henter, S.~L.
  Maguer, Z.~Malisz, Éva Székely, C.~Tånnander, and J.~Voße, ``Speech
  synthesis evaluation — state-of-the-art assessment and suggestion for a
  novel research program,'' {\em Proc. 10th ISCA Speech Synthesis Workshop},
  2019.

\end{thebibliography}
\end{document}